\documentclass{article}
\PassOptionsToPackage{numbers, compress}{natbib}

\usepackage[preprint]{neurips_2025} 
\usepackage{float} 
\usepackage{multirow}  
\usepackage{booktabs}
\usepackage{subcaption} 

\usepackage{graphicx}
\usepackage[utf8]{inputenc} 
\usepackage[T1]{fontenc}    
\usepackage{hyperref}       
\usepackage{url}            
\usepackage{booktabs}       
\usepackage{amsfonts}       
\usepackage{nicefrac}       
\usepackage{microtype}      
\usepackage{xcolor}         
\usepackage{booktabs}
\usepackage{subcaption}
\usepackage{amsmath}
\usepackage{bm}

\title{ELITE: Embedding-Less retrieval with Iterative Text Exploration}

\author{%
  Zhangyu Wang\thanks{Co-first author, order decided with a coin flip.} \\
  University of Southern California \\
  \texttt{zhangyuw@usc.edu}
  \And
  \textbf{Siyuan Gao}\footnotemark[1] \\
  Jilin University \\
  \texttt{gaosy2122@mails.jlu.edu.cn}
  \And
  Rong Zhou \\
  National Supercomputing \\Center in Shenzhen \\
  \texttt{zhourong0925@gmail.com}
  \And
  \textbf{Hao Wang}\thanks{Co-corresponding author.} \\
  Wuhan University \\
  \texttt{wanghao.cs@whu.edu.cn}
  \And
  Li Ning\thanks{Corresponding author.} \\
  Stellaris AI Limited \\
  \texttt{lining@stellaris-ai.com}
}

\begin{document}
\maketitle

\begin{abstract}
Large Language Models (LLMs) have achieved impressive progress in natural language processing, but their limited ability to retain long-term context constrains performance on document-level or multi-turn tasks. Retrieval-Augmented Generation (RAG) mitigates this by retrieving relevant information from an external corpus. However, existing RAG systems often rely on embedding-based retrieval trained on corpus-level semantic similarity, which can lead to retrieving content that is semantically similar in form but misaligned with the question's true intent. Furthermore, recent RAG variants construct graph- or hierarchy-based structures to improve retrieval accuracy, resulting in significant computation and storage overhead. In this paper, we propose an embedding-free retrieval framework. Our method leverages the logical inferencing ability of LLMs in retrieval using iterative search space refinement guided by our novel importance measure and extend our retrieval results with logically related information without explicit graph construction. Experiments on long-context QA benchmarks, including NovelQA and Marathon, show that our approach outperforms strong baselines while reducing storage and runtime by over an order of magnitude.\footnote{Code available at \href{https://github.com/tjzvbokbnft/ELITE-Embedding-Less-retrieval-with-Iterative-Text-Exploration}{our GitHub repository}} 
\end{abstract}
\section{Introduction}
\label{section:intro}

Recent advancements in Large Language Models (LLMs) have revolutionized natural language processing, enabling sophisticated text generation and applications across diverse domains.
Despite their impressive capabilities, LLMs still face significant limitations in maintaining context over extended conversations or lengthy documents, primarily due to the lack of built-in memory mechanisms. A naive approach of feeding all past context into the model for every query is usually thought to be computationally expensive and impractical for real-world deployment.
\emph{Retrieval-Augmented Generation} (RAG) has emerged as a prominent framework to overcome these limitations by enabling access to external knowledge through explicit retrieval mechanisms \citep{lewis2021retrievalaugmentedgenerationknowledgeintensivenlp}.

A typical RAG system consists of two phases: an \textbf{offline phase} for indexing and preparation, and an \textbf{online phase} for retrieval and generation. 

\textbf{Offline Phase: Indexing and Preparation} \\
In this phase, source documents are split into manageable chunks, and a certain form of index is assigned to each chunk that will be compared to find related information to answer the query in the later retrieval stage. Usually, the index is the vector embedding of the chunk. After assigning indexes, for some methods, an additional summarization or knowledge graph construction step will be performed to compress chunk information or populate the textual database with richer relational information. Notably, offline phase processing time could vary greatly between different methods, and for a real-life application that values user experience, taking a long time to pre-process an input text before Question Answering regarding the text is undesirable.

\textbf{Online Phase: Retrieval and Generation} \\
Online phase, on the other hand, is executed in real time when a query is received. It includes two sub-stages:
\textbf{Retrieval Stage:} The system encodes the query and searches for relevant document chunks in the indexed corpus. Nearly all existing RAG systems rely on retrievers using embedding similarity for this step. Some methods sought to improve retrieval accuracy through query expansion with LLM or using a fine-tuned embedding model designed to encode a query and the answers differently, targeting specifically for the retrieval task~\citep{sturua2024jinaembeddingsv3multilingualembeddingstask}.
\textbf{Generation Stage:} Retrieved content is combined with the query input and fed into the LLM to generate the final response. The overall performance at this stage typically hinges on both the quality of retrieved information and the generation/inferential capability of the LLM.

Currently, a critical component in the RAG pipeline is the embedding models in both Offline and Online phases. Embedding-based dense retrieval is the dominant approach in RAG due to its assumed ability to capture latent semantic relationships and generalize across lexically divergent queries and documents.

However, embedding models might not be able to capture the complex semantic relationship between question and answer pairs and tend to retrieve syntactically similar sentences rather than those that answer the question, resulting in poor retrieval accuracy. For instance, a query like \textit{``how many times has someone done something?''} may retrieve dialogues containing \textit{``how many times you have done this?''}, rather than factual descriptions of someone’s actual behavior. Our experiments in Section \ref{sec:emb_lim} show that relying on embedding similarity could result in undesired performance in some cases across various embedding models, regardless of whether the embedding model is fine-tuned for the retrieval task or not. Besides, embedding models are black boxes and have poor interpretability, such that one cannot flexibly modify their retrieval policy when using a specific embedding model that resulted in poor retrieval performance on some queries. 


In addition, embedding-based approaches suffer from other practical drawbacks. Many recent systems construct complex graph or hierarchical structures at the offline phase using LLMs and embeddings, introducing significant computation time and additional storage overhead.

\textbf{In terms of computation time}, when documents are long and user queries are sparse, the average query latency increases substantially. This leads to poor responsiveness and diminished user experience. While such overhead might be acceptable in high-query-volume applications, it becomes inefficient in real-world scenarios involving only a few queries per document. In the longest context interval ($>$2e6), MiniRAG\citep{fan2025miniragextremelysimpleretrievalaugmented} incurs a total runtime that is \textbf{1035.9$\times$} higher than our method, while RAPTOR\citep{sarthi2024raptorrecursiveabstractiveprocessing} takes \textbf{344.2$\times$} longer-highlighting a major efficiency gap.

\textbf{In terms of storage}, RAPTOR and MiniRAG produce summarized or extracted data that greatly exceeds the original input size. This includes embedding vectors, graph structures, and summary texts. Specifically, RAPTOR incurs a \textbf{19.6$\times$} increase in data volume, while MiniRAG results in a \textbf{14.7$\times$} increase.

In summary, current embedding-based RAG methods face several limitations:
\begin{enumerate}
    \item Rely on black-box embedding models for indexing and retrieval, which can result in undesirable and inflexible performance;
    \item Utilized complex data structures in offline phase to enhance performance, which introduce additional computation time and storage overhead during indexing;
    \item Complex methods lead to substantial storage expansion, with storage requirements often exceeding the original text size by over 10 times.
\end{enumerate}

To overcome these challenges, we propose an embedding-free retrieval framework that leverages LLMs’ native reasoning capabilities rather than vector similarity, which not only enables us to adjust our retrieval policy in real time based on the retrieved results to perform more flexible and accurate retrieval, but also has minimal storage and time complexity. Also, although our method does not rely on any explicit graph construction, we still manage to exploit relational information with our special extension method.

Extensive experiments on popular long-context QA datasets, such as \textit{NovelQA} and \textit{Marathon}, demonstrate that our method consistently outperforms embedding-based baselines in both effectiveness and efficiency.\\

In summary, our key contributions are summarized below:
\begin{enumerate}
\item \textbf{Fast and Lightweight.}  
Our method eliminates reliance on embedding models and dense indexing by using LLM reasoning directly. It requires no extra storage, unlike MiniRAG and Raptor which incur up to \textbf{10.6$\times$–19.6$\times$} overhead, and delivers rapid responses. On long documents ($>$2M tokens), it reduces preparation time from \textbf{19,337.6s} and \textbf{6,423.2s} to just \textbf{0.085s}, with total runtime cut to \textbf{18.7s}.

\item \textbf{High Retrieval Accuracy.}
Without relying on embeddings or knowledge graphs, our method achieves 71.27\% on NovelQA and 68.46\% on Marathon using LLaMA3.1:70b, ranking \textbf{2nd} overall on both benchmarks.
The top-performing models are \textbf{GPT-4-0125-preview} (71.80\%) for NovelQA and \textbf{GPT-4} (78.59\%) for Marathon.

\item \textbf{Robust Generalization.}  
Our method consistently improves with model scale across both NovelQA and Marathon. From 1b to 70b, it shows stable gains, outperforming all baseline methods at every scale. This highlights strong scalability, architecture independence, and robustness across diverse long-context QA tasks.

\end{enumerate}

\section{Related Work}
\textbf{Embedding Models.} The concept of word embeddings was first introduced by Bengio et al.~\citep{10.5555/944919.944966} in a neural probabilistic language model that jointly learned word representations and a language model. This inspired static embeddings like Word2Vec~\citep{mikolov2013efficientestimationwordrepresentations} and GloVe~\citep{pennington-etal-2014-glove}, which map each word to a fixed vector based on co-occurrence statistics. Contextual embeddings, including ELMo~\citep{peters2018deepcontextualizedwordrepresentations} and BERT~\citep{devlin-etal-2019-bert}, later captured dynamic, sentence-aware representations. Sentence-level encoders such as SBERT~\citep{reimers-gurevych-2019-sentence} improved embedding quality for semantic similarity and retrieval. Recently, scalable embedding models like Nomic's Text Embeddings~\citep{nussbaum2025nomicembedtrainingreproducible}, Nvidia's NVEmbed~\citep{lee2025nvembedimprovedtechniquestraining}, and Jina Embeddings~\citep{sturua2024jinaembeddingsv3multilingualembeddingstask} have emerged, optimized for downstream tasks and large-scale vector search.

\textbf{Retrieval-Augmented Generation.} Retrieval-Augmented Generation (RAG) was first proposed by Lewis et al.~\citep{lewis2021retrievalaugmentedgenerationknowledgeintensivenlp} as a framework combining dense retrievers (e.g., DPR) with sequence-to-sequence generators (e.g., BART). It significantly improved open-domain QA, but suffered from noisy retrieval and lacked long-context modeling. Follow-up work explored better fusion strategies and end-to-end training, such as FiD~\citep{izacard2021leveragingpassageretrievalgenerative} and REALM~\citep{guu2020realmretrievalaugmentedlanguagemodel}. Recent models such as Atlas~\citep{izacard2022atlasfewshotlearningretrieval} and RePlug~\citep{shi2023replugretrievalaugmentedblackboxlanguage} propose decoupling retriever training from generators, enabling modularity and improving zero-shot generalization. ColBERT-RAG~\citep{khattab2020colbertefficienteffectivepassage} replaces DPR with late-interaction retrievers, allowing finer-grained token-level relevance matching for dense passage retrieval. Other approaches like Memory RAG~\citep{qian2025memoragboostinglongcontext} explores explicit memory structures or token-level caches to better retain long-range factual knowledge. Graph-based variants (Graph-RAG~\citep{edge2025localglobalgraphrag}, Path-RAG~\citep{chen2025pathragpruninggraphbasedretrieval}) introduced explicit structural guidance for multi-hop reasoning but incurred substantial overhead. More recent advances, such as MiniRAG~\citep{fan2025miniragextremelysimpleretrievalaugmented} and Raptor~\citep{sarthi2024raptorrecursiveabstractiveprocessing}, simplify retrieval pipelines by replacing static, large-scale graphs with lighter-weight structures or recursive chunk hierarchies.

\section{Limitations of Embedding}
\label{sec:emb_lim}

\begin{table}[htbp]
\centering
\scriptsize
\setlength{\tabcolsep}{4pt}
\renewcommand{\arraystretch}{0.9}
\begin{tabular}{|l|c|c|c|c|c|}
\hline
\textbf{Model} & \textbf{Category} & \textbf{Q ANS} & \textbf{REL CLUE} & \textbf{IRR CLUE} & \textbf{RANDOM} \\
\hline
\multirow{3}{*}{Mxbai-embed-large} 
& Character-sh & 0.4326 & 0.5441 & \textbf{0.5978} & 0.3734 \\
& Relat-mh     & \textbf{0.6890} & 0.5323 & 0.6010 & 0.3395 \\
& Span-mh      & 0.4778 & 0.4242 & \textbf{0.5596} & 0.3184 \\
\hline
\multirow{3}{*}{NV-Embed-v1\cite{lee2025nvembedimprovedtechniquestraining}} 
& Character-sh & 0.3333 & 0.2786 & \textbf{0.4397} & 0.2952 \\
& Relat-mh     & \textbf{0.5487} & 0.3192 & 0.4842 & 0.3078 \\
& Span-mh      & 0.4247 & 0.2196 & \textbf{0.4696} & 0.2436 \\
\hline
\multirow{3}{*}{Jina-v3-text-matching\cite{sturua2024jinaembeddingsv3multilingualembeddingstask}} 
& Character-sh & 0.4855 & 0.5947 & \textbf{0.6584} & 0.4492 \\
& Relat-mh     & \textbf{0.7330} & 0.6043 & 0.6631 & 0.4185 \\
& Span-mh      & 0.5488 & 0.4843 & \textbf{0.6332} & 0.3854 \\
\hline
\multirow{3}{*}{Jina-v3-retrieval\cite{sturua2024jinaembeddingsv3multilingualembeddingstask}} 
& Character-sh & 0.0002 & \textbf{0.3210} & 0.2205 & 0.0837 \\
& Relat-mh     & \textbf{0.4050} & 0.3483 & 0.2417 & 0.0410 \\
& Span-mh      & 0.1744 & 0.1828 & \textbf{0.2008} & 0.0042 \\
\hline
\end{tabular}
\caption{Performance of common embedding models on NovelQA tasks. “Character”, “Relat”, and “Span” indicate question types; “s” denotes single-hop reasoning, “m” indicates multi-hop.}
\label{tab:embedding_performance}
\end{table}
We evaluated common embedding model retrieval performance through a four-part analysis: (1) Query-Answer embedding similarity using samples from NovelQA, (2) similarity between queries and LLM-generated relevant clues designed to answer the query with a specific answer, (3) similarity between queries and LLM-generated irrelevant clues that mimic the syntactic structure of the query without providing answer-relevant information, and (4) similarity between queries and randomly sampled passage sentences. As shown in Table \ref{tab:embedding_performance}, all models except Jina-v3-retrieval, which is the best SoTA embedding model that embeds queries and answers separately and specifically fine-tuned for retrieval tasks, shows a preference toward Irrelevant clues over actual clues, and are dangerously close to random sampling in some categories. Even though using different embedding method for query and answers in Jina-v3-retrieval shows promising results in most cases, there are still categories like Span-mh where it favors syntactical similarity over the actual relationship between the question and answers. Given the observations that all embedding models have certain cases where they could fail to retrieve target information, the challenge with using embedding in retrieval is thus: how do we know at test time, when we are not given these categories, that our embedder might not perform ideally? And, even if we know, how can we modify the black boxed model's retrieval policy to make it retrieve something else at test time? These challenges serve as constraints that limit the upper boundary for embedding-based methods, and are quite difficult, if not impossible, to resolve. Using LLMs to extend the query might not be a solution either, as we cannot add extra information into the query that always maps it closer to the answer in the embedding space unless the LLM explicitly knows the answer. Not to mention that query extension could introduce noise and even more uncertainty in our retrieval process, making it even harder to change the retrieval policy when needed.
\section{Methodology}
\begin{figure*}
    \centering
    \begin{subfigure}{1\textwidth}
        \includegraphics[width=\linewidth]{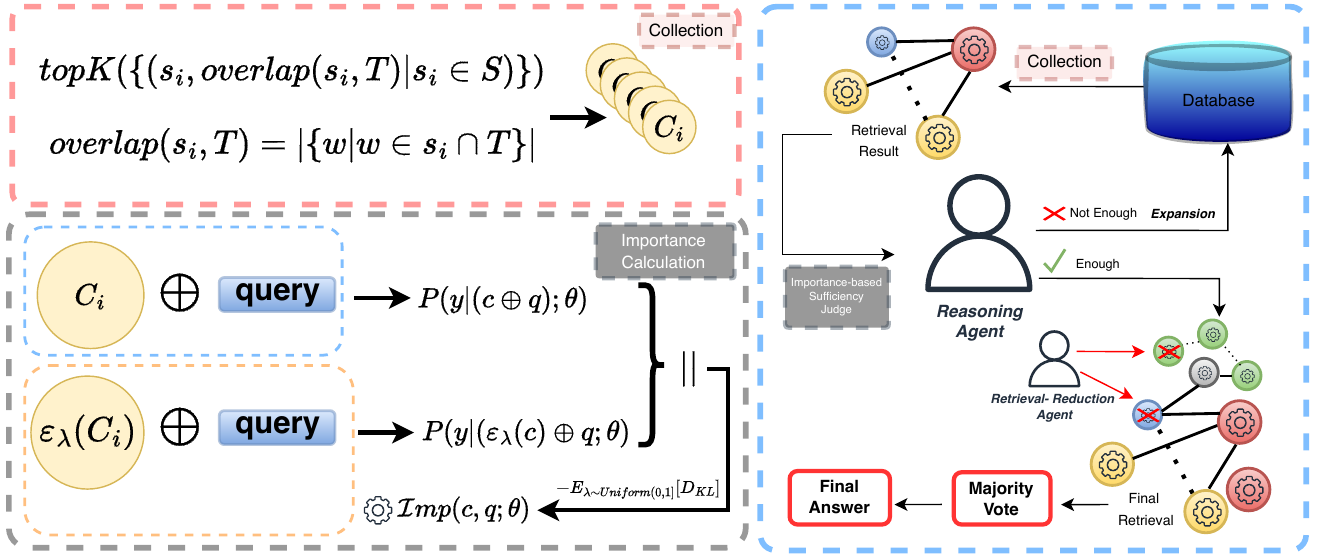}
        
    \end{subfigure}%
    \caption{algorithm flowchart}
    \label{fig:flowchart}
\end{figure*}

In this section, we present our method, which transforms the Question-Answer matching problem into an explicit search and filter operation that leverages LMs' inferential and language capability instead of using vector embeddings to produce rapid and accurate responses. 

To start with, our method draws inspiration from human cognitive processes employed during long-text question answering tasks. When confronted with extensive textual material and specific queries, individuals typically engage in a sequence of structured cognitive operations: first, the identification and generation of phrases and words of interest based on the question(Generation); followed by the data collection through retrieving the part of source material containing the words or phrases of interest(Collection); subsequently, the evaluation of the sufficiency of retrieved source data(Evaluation); the iterative expansion of search scope when retrieved information proves insufficient(Exploration); and lastly, trimming that deletes irrelevant information and reorganize evidence to produce a coherent response(Organizing). 

Drawing inspiration from this cognitive paradigm, we formulate a methodology that emulates these natural information processing strategies. Our method can be described with four main stages: Agent-guided texts of interest generation(Generation), data comparison and retrieval(Collection), agent-oriented subjective information sufficiency decision aided with our objective ``importance'' metrics(Evaluation/Importance-based Sufficiency Judge), agent-driven breadth-wise and depth-wise extension to modify our retrieval policy(Exploration), and finally, Agent-based evidence trimming and response generation(Organizing). These general stages of our method are depicted in Figure \ref{fig:flowchart}. 

\subsection{Generation and Collection}
During our experimental analysis of the novelQA dataset, we realized that questions in the ``times'' category-which require enumerating occurrences of specific events, words, or phrases-significantly challenge our model's counting capabilities. Due to the lack of accurate counting ability in our base model, we observed diminished accuracy specifically for this question type. However, such quantification queries constitute an essential component of long-context comprehension systems, and proficiency in addressing such questions is imperative for the deployment of effective question answering frameworks in practical applications. Benefiting from the flexibility of our method, during our Generation phase, we could first prompt the model to identify whether the question involves accurate counting. Once a problem is identified as a ``counting'' type, it would be solved separately from other questions; more specifically, we will prompt the model to generate only one phrase or word of interest, then generate all tenses(past, present continuing) or forms(plural, etc.) of it if applicable. Then we will count the total number of occurrences of the list of words/phrases in different tenses/forms within the passage and skip all later phases and directly report the result. 

For non-counting questions, our methodology implements a relevance-based lexical retrieval approach in the Generation and Collection steps. The agent is first prompted to generate a set of contextually pertinent phrases or lexical units derived from either the query itself or concepts essential to its resolution. Next, we employ a simple but efficient word-overlap algorithm to identify and extract sentences within the passage that contain these target terms. Retrieved segments are ranked according to total lexical overlap in descending order and top K sentences are passed to the next step. The Collection step can be described by the following equations: 
\begin{equation}
  \label{eq:overlap_score}
  overlap(s_i, T) = |\{w | w \in s_i \cap T\}| 
\end{equation}
\begin{equation}
  \label{eq:collection}
  R = topK(\{(s_i, overlap(s_i, T) | s_i \in S)\})
\end{equation}

Where T is the target terms, S is the set of all sentences in the passage, and R is the retrieved results. 
To preserve contextual integrity, we implement a naive symmetric contextual expansion procedure for each sentence that incorporates five sentences from both preceding and subsequent textual regions for each identified sentence. These expanded sentences(chunks) will be passed to the evaluation phase.

\subsection{Evaluation/Importance-based Sufficiency Judge}
Following the retrieval of relevant text chunks, we task our base model with assessing whether the retrieved information suffices to answer the given query. However, this assessment faces two significant challenges: (1) the model lacks intrinsic knowledge about how influential specific chunks are in its reasoning process, and (2) the evaluation task is inherently subjective without clear objective criteria guidance. To address these limitations, we propose a novel objective metric-the ``importance'' score-that quantifies the contribution of retrieved text segments to the model's decision-making process.

We formally define the ``importance'' of a chunk $c$ to a query $q$ as the expected KL divergence between the posterior probability distributions of the model's output, comparing responses generated with the original text versus those generated when the chunk is perturbed with varying levels of noise:
\begin{equation}
    \label{eq:importance}
    \mathcal{I}mp(c, q; \theta) = -E_{\lambda \sim Uniform(0, 1]}[D_{KL}(P(y | (c \oplus q); \theta) || P(y | (\varepsilon_\lambda(c) \oplus q); \theta))]
\end{equation}
Where c is the collected chunks, q is the query text, $\lambda$ denotes the noise level (probability of noise added to each character), $\oplus$ denotes concatenation operation, and $\theta$ is the language model. Given the computational challenges in directly calculating probability distributions over the model's text output space, we approximate this expectation through:
\begin{equation}
    \label{eq:importance_simp}
    \mathcal{I}mp(c, q; \theta) = 1 - AvgSim(\theta (c \oplus q), \theta(\varepsilon_\lambda(c) \oplus q))
\end{equation}
This approximation is justified under the assumption that the output distributions maintain the same functional form with variance determined by the fixed temperature parameter in the model configuration. Consequently, the mean becomes the primary determinant in the KL divergence calculation, with the original expectation monotonically related to the similarity between the means. We implement a Monte Carlo method to estimate this relationship by randomly sampling multiple noise levels and computing the average similarity between outputs from original and noise-perturbed inputs. In our actual implementation, we used cosine similarity between embedded result texts to calculate the similarity, which works because the resulting answer is expected to have highly similar format and semantic information unless the answer to the question has changed or the reasoning logic has changed due to noise added to the chunk, which implies that the chunk is highly influential to the model's answer. During implementation, we added noise to the strings using the package from \url{https://github.com/dleemiller/string-noise} by Lee Miller et al., which adds more natural and diverse noise to the string than simply swapping random characters.

After computing these objective ``importance'' scores, we incorporate them as supplementary information alongside the task prompt and retrieved chunks to guide the model's assessment. Importantly, we do not rely solely on the importance score to determine information sufficiency, but rather leverage it as additional evidence to support the model's reasoning. This approach acknowledges that while the importance score quantifies a chunk's influence on the model's output, it does not necessarily indicate whether the information is sufficient to comprehensively answer the query.

\subsection{Iterative Text Exploration and Organizing}
Once the model identifies the collected chunks as insufficient to answer the query, Exploration and Evaluation steps will be performed iteratively to check for sufficiency and extend until either the model considers collected chunks to be sufficient or an upper threshold is hit. Through these two steps we are able to dynamically modify our retrieval policy/search scope based on previous retrieval results and policy.

Recent successes in applying graph-based methods shows the ability of graphs in enhancing the retrieval accuracy and providing background contexts, but explicitly constructing relational graphs can be highly computationally expensive. To include related background information and effectively extend our search space with minimal computational cost, we designed our breadth and depth wise implicit graph extension. During our extension step, we employ a two-phase approach that emulates graph-based information propagation without explicitly constructing the computationally expensive graph structures. 

First, we re-generate an expanded list of search words/phrases based on previous retrieval results to dynamically broaden our search space, effectively traversing the implicit knowledge graph in a breadth-first manner. This process captures semantically adjacent concepts that may not have been included in the initial retrieval. Subsequently, we generate a secondary list of search words/phrases derived from the initially retrieved chunks, enabling a depth-first exploration of the implicit knowledge graph by identifying logically or semantically relevant contexts for our retrieved information.

This methodology can be formalized as traversing an implicit knowledge graph $G = (V, E)$, where vertices $V$ represent concepts and edges $E$ represent semantic or logical relationships. Our breadth-wise extension corresponds to expanding and identifying new discrete base vertices $v_i \in V_i \cup N_d(v_p)$ where $N_d(v_p)$ represents the logical or semantical neighbor of vertices from previous list $v_p$, and $V_i$ denotes the newly identified nodes generated by model's inference the needed information based on the previous list, query text, and retrieval results. Concurrently, our depth-wise extension identifies vertices $v_j \in N_d(v_r)$, where $N_d(v_r)$ represents the depth-wise neighborhood of retrieved nodes $v_r$.

The implicit nature of our approach obviates the need for explicit graph construction and traversal algorithms, which typically incur $O(|V| + |E|)$ computational complexity and involve frequent inference of the relationships for all edges with the LM. Instead, by leveraging the generative capabilities of large language models to identify semantically related terms, we achieve comparable information propagation with significantly reduced computational overhead, approximating the benefits of graph-based methods while maintaining efficiency comparable to traditional retrieval approaches.

Lastly, when either the upper threshold is hit or the model decided the collected chunks are sufficient, we will prompt the model to filter out information irrelevant to answering the question in the collected chunks to minimize disturbance and organize the information.

\section{Experiments}
\paragraph{Benchmarks.}
We evaluated our method on two long-context multiple-choice QA benchmarks-- \textbf{NovelQA}\cite{wang2024novelqabenchmarkingquestionanswering} and \textbf{Marathon}\cite{zhang2024marathonracerealmlong}. 

\textbf{NovelQA}~\cite{wang2024novelqabenchmarkingquestionanswering} is a long-context QA benchmark with 2,305 multiple-choice questions from 88 novels, spanning aspects such as plot, character, and setting. Contexts exceed 200,000 tokens, demanding advanced reasoning. We evaluate models by accuracy on these multiple-choice tasks.

\textbf{Marathon}~\cite{zhang2024marathonracerealmlong} is a long-context QA benchmark with 1,530 multiple-choice questions covering six task types, including reasoning and retrieval. Each question has one correct answer and three verified distractors. Contexts average 100K characters, with some exceeding 260K, requiring deep reasoning under extreme length constraints.\\

\textbf{Baselines. }
\textbf{Comparison to Existing RAG Methods. }We compare our method with several cost-efficient RAG baselines on the full benchmark datasets (no sampling).
\textbf{Internal}: Relies solely on the model’s pre-trained knowledge. Included to assess how much external retrieval improves performance, especially since NovelQA includes well-known literary works.
\textbf{Vanilla}: Uses the model’s long-context ability by directly appending documents into the input without retrieval modules.
\textbf{MiniRAG}~\cite{fan2025miniragextremelysimpleretrievalaugmented}: Builds a lightweight semantic-aware graph for retrieval, enabling small models to perform competitively with low memory and storage overhead.
\textbf{RAPTOR}~\cite{sarthi2024raptorrecursiveabstractiveprocessing}: Applies recursive, tree-structured summarization for hierarchical retrieval, enhancing long-context reasoning with fewer tokens
\textbf{Comparison with SOTA Language Models. }We also compare our proposed method with state-of-the-art (SoTA) language models on both the NovelQA and Marathon benchmarks. The results for the SoTA models on Table~\ref{tab:sidebyside} are obtained from the official leaderboards and corresponding publications for NovelQA and Marathon.\\

\textbf{Configurations. }We evaluated our method using LLaMA-3.1 (8B, 70B) and LLaMA-3.2 (1B, 3B). All models except 70b were tested on a single RTX 4070 Ti SUPER GPU with 16GB of VRAM. The 70b model was evaluated on a Mac Studio with an M2 Ultra chip and 192 GB unified memory.\\
\textbf{Experimental Setting. }For our method, we set \texttt{recall\_index=25}, \texttt{neighbor\_num=2}, \texttt{deep\_search\_index=5}, \texttt{deep\_search\_num=10}, \texttt{voter\_num=10} and \texttt{iter\_max=5}. For all baselines, we strictly follow their original settings.

\subsection{Results}

\begin{table}[h]
    \centering
    \scriptsize
    \renewcommand{\arraystretch}{1.1}
    
    \begin{subtable}[t]{0.48\linewidth}
        \centering
        \begin{tabular}{lcccc}
            \toprule
            & \multicolumn{2}{c}{LLaMA 3.2} & \multicolumn{2}{c}{LLaMA 3.1} \\
            Method & 1b & 3b & 8b & 70b \\
            \midrule
            internal & 24.09 & 30.56 & 26.37 & 38.04 \\
            vanilla  & 24.22 & 36.57 & 56.02 & 44.55 \\
            MiniRAG  & 15.94 & 41.79 & 45.34 & 50.34 \\
            Raptor   & 14.81 & 50.37 & 52.56 & 58.96 \\
            Ours     & \textbf{29.57} & \textbf{55.54} & \textbf{65.18} & \textbf{71.27} \\
            \bottomrule
        \end{tabular}
        \caption{NovelQA ACC (LLaMA)}
    \end{subtable}%
    \hfill
    \begin{subtable}[t]{0.48\linewidth}
        \centering
        \begin{tabular}{lcccc}
            \toprule
            & \multicolumn{2}{c}{LLaMA 3.2} & \multicolumn{2}{c}{LLaMA 3.1} \\
            Method & 1b & 3b & 8b & 70b \\
            \midrule
            internal & 20.42 & 33.61 & 36.21 & 43.74 \\
            vanilla  & 22.29 & 54.84 & 61.34 & 66.29 \\
            MiniRAG  & 21.32 & 37.30 & 43.56 & 46.88 \\
            Raptor   & \textbf{25.44} & 49.50 & 56.46 & 65.17 \\
            Ours     & 25.21 & \textbf{55.67} & \textbf{61.45} & \textbf{68.46} \\
            \bottomrule
        \end{tabular}
        \caption{Marathon ACC (LLaMA)}
    \end{subtable}

    \vspace{0.5em}

    \begin{subtable}[t]{0.48\linewidth}
        \centering
        \begin{tabular}{lc}
            \toprule
            Model & ACC \\
            \midrule
            GPT-3.5+RAG Langchain & 56.94 \\
            Claude-v2:1 vanilla   & 66.84 \\
            GPT-4+RAG Langchain   & 67.89 \\
            Claude-3-Sonnet       & 71.11 \\
            GPT-4-0125-preview    & 71.80 \\
            Human performance     & 97.00 \\
            \bottomrule
        \end{tabular}
        \caption{NovelQA ACC (Other LLMs)}
    \end{subtable}%
    \hfill
    \begin{subtable}[t]{0.48\linewidth}
        \centering
        \begin{tabular}{lc}
            \toprule
            Model & ACC \\
            \midrule
            Mistral-7B+naiveRAG+OpenAIEmbed & 50.18 \\
            Qwen-14B vanilla                & 39.27 \\
            Qwen-14B+RAG+JinaEmbed          & 58.12 \\
            Yi-chat-34B vanilla             & 55.91 \\
            Yi-chat-34B+RAG+JinaEmbed       & 63.81 \\
            GPT-4 vanilla                   & 78.59 \\
            \bottomrule
        \end{tabular}
        \caption{Marathon ACC (Other LLMs)}
    \end{subtable}

    \caption{Performance Comparison on NovelQA and Marathon}
    \label{tab:sidebyside}
\end{table}

Table~\ref{tab:sidebyside} compares QA accuracy across retrieval strategies on NovelQA and Marathon.\\
\textbf{NovelQA.} Our method consistently outperforms baselines across all LLaMA scales. At the 70b scale, it achieves 71.27\%, surpassing MiniRAG (50.34\%) and Raptor (58.96\%). Even with only 8b parameters, it exceeds GPT-3.5+RAG (56.94\%) and approaches Claude-v2 (66.84\%). These results demonstrate strong generalization under limited capacity and near parity with proprietary systems.\\
\textbf{Marathon.} Our method shows stable gains across scales, overtaking Raptor from 3b upward. At 70b, it reaches 68.46\%, outperforming Raptor (65.17\%) and MiniRAG (46.88\%). The trend indicates robust scalability and long-context reasoning, approaching GPT-4 (78.59\%) with significantly fewer resources.

\subsubsection{Response Time Analysis}

\begin{figure}[h]
    \centering
    \begin{minipage}{0.40\textwidth}
        \begin{subfigure}{\linewidth}
            \includegraphics[width=\linewidth]{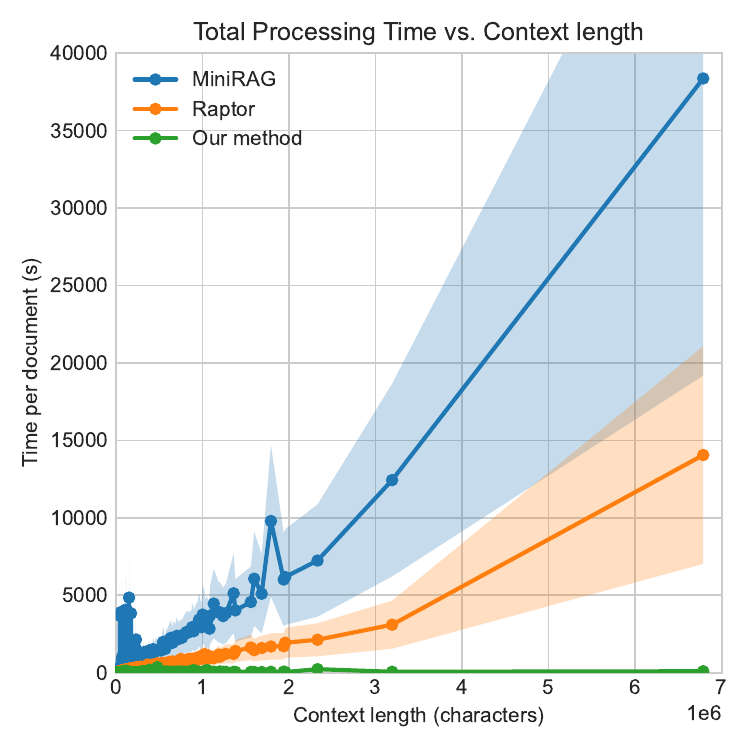}
            \caption{Total Processing Time}
            \label{fig:time_total}
        \end{subfigure}
        
    \end{minipage}%
    \begin{minipage}{0.40\textwidth}
        \begin{subfigure}{\linewidth}
            \includegraphics[width=\linewidth]{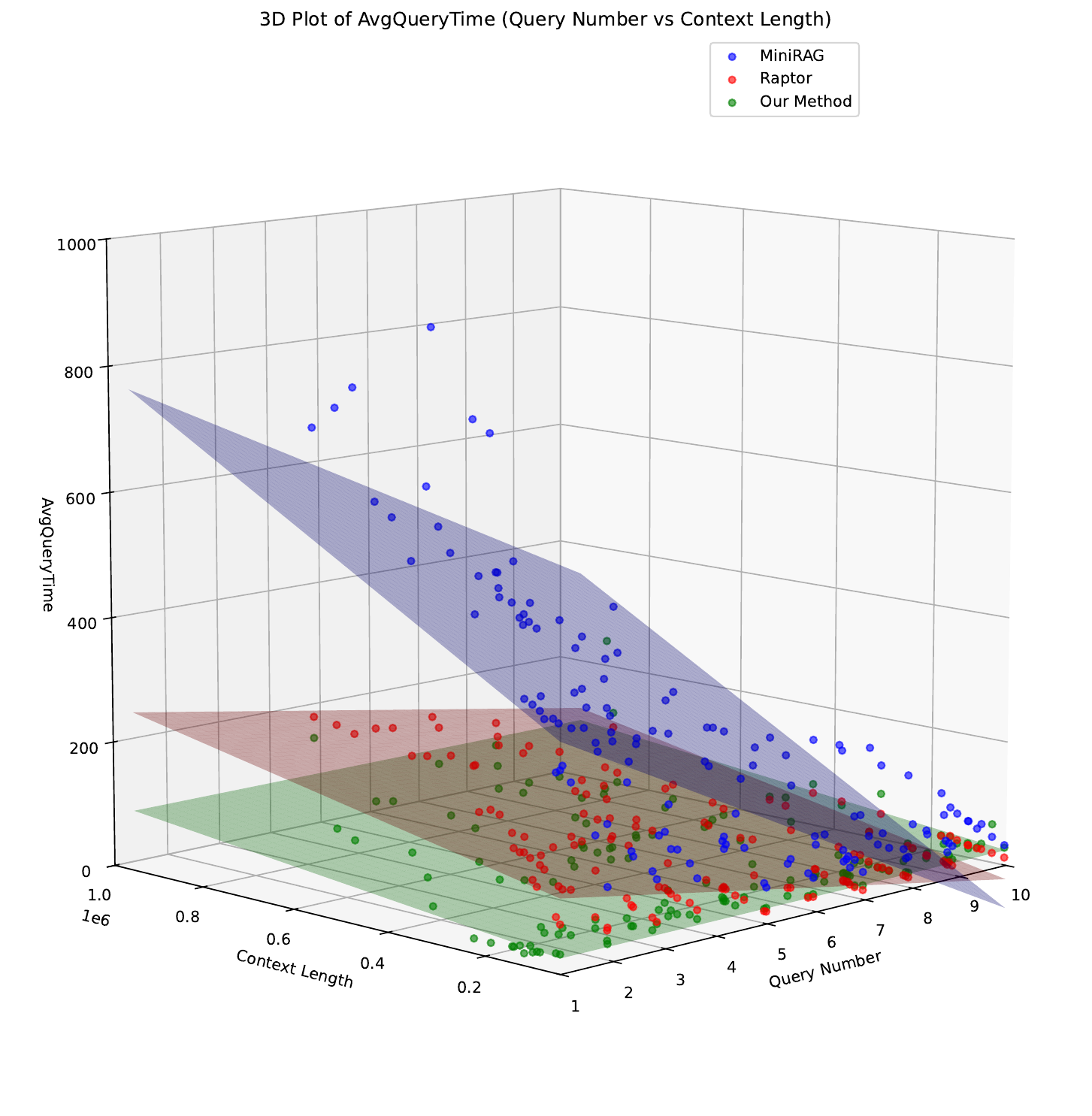}
            \caption{Average Query Time (3D)}
            \label{fig:avg_query}
        \end{subfigure}
    \end{minipage}

    \caption{Breakdown of total time consumption across different stages.}
    \label{fig:response_time_breakdown}
\end{figure}

\begin{table}[htbp]
    \centering
    
    \resizebox{\textwidth}{!}{%
    \begin{tabular}{
        l                
        r                
        r r r r          
        r r r r          
        r r r r          
    }
    \toprule
    \textbf{Interval} & \textbf{Avg Context Length} &
    \multicolumn{4}{c}{\textbf{MiniRAG}} &
    \multicolumn{4}{c}{\textbf{Raptor}}  &
    \multicolumn{4}{c}{\textbf{Our Method}} \\
    \cmidrule(lr){3-6}\cmidrule(lr){7-10}\cmidrule(lr){11-14}
                      & &
        Prep. & Retr. & Gen. & Total &
        Prep. & Retr. & Gen. & Total &
        Prep. & Retr. & Gen. & Total \\
    \midrule
    $<$1e5      &  79,028.343 & 348.630 & 6.564 & 1.511 & 356.705 & 108.741 & 0.031 & 4.600 & 113.372 & 0.004 & 0.774 & 2.725 & \textbf{3.503} \\
    (1e5,5e5)   & 170,635.733 & 648.515 & 7.757 & 1.721 & 657.993 & 208.254 & 0.042 & 4.906 & 213.203 & 0.007 & 0.975 & 2.984 & \textbf{3.966} \\
    (5e5,1e6)   & 752,523.312 & 2,390.854 & 6.193 & 4.202 & 2,401.249 & 759.523 & 0.116 & 6.475 & 766.114 & 0.017 & 1.684 & 6.485 & \textbf{8.186} \\
    (1e6, 2e6)  & 1,346,647.368 & 4,564.582 & 6.520 & 4.189 & 4,575.291 & 1,268.484 & 0.123 & 6.321 & 1,274.929 & 0.030 & 2.451 & 6.523 & \textbf{9.004} \\
    $>$2e6      & 4,103,032.667 & 19,337.624 & 6.849 & 4.518 & 19,348.990 & 6,423.175 & 0.181 & 6.344 & 6,429.700 & 0.085 & 6.047 & 12.546 & \textbf{18.678} \\
    \bottomrule
    \end{tabular}}
    \caption{Average runtime of different methods across context-length intervals and processing phases (in seconds)}
    \label{tab:runtime_full_en}
\end{table}

We evaluate response time using LLaMA-3.1:8b, dividing the QA process into \textbf{offline} and \textbf{online} phases. The \textbf{offline phase} includes \textit{preparation time}, covering graph construction, chunking, and embedding-based indexing. The \textbf{online phase} comprises \textit{retrieval time}, the latency of fetching relevant content, and \textit{generation time}, which measures the duration from receiving the context to producing the final answer.\\

Table~\ref{tab:runtime_full_en} presents a detailed breakdown of the average runtime for MiniRAG, Raptor, and our method across different context-length intervals and processing phases. Across all intervals, our method consistently demonstrates the lowest total latency, with total processing time ranging from only 3.5 to 18.7 seconds, while MiniRAG and Raptor exhibit significant increases up to 19,349.0 and 6,429.7 seconds respectively-primarily due to the overhead in the preparation phase.

Preparation time dominates the total cost for both MiniRAG and Raptor, particularly as the context length increases, reaching 19,337.6 seconds and 6,423.2 seconds in the $>$2e6 interval. In contrast, our method achieves near-zero preparation overhead (e.g., 0.085 seconds), as it does not require graph construction or embedding-based indexing. Retrieval time remains low and stable across all methods, while generation time increases moderately with context size but remains manageable.

To better capture per-query efficiency, we define the Average Query Time as:
\begin{equation}
\text{AvgQueryTime} = \frac{T_{\text{prepare}} + N \cdot (T_{\text{retrieve}} + T_{\text{gen}})}{N}
\end{equation}
where $T_{\text{prepare}}$ is the one-time preprocessing cost, $T_{\text{retrieve}}$ and $T_{\text{gen}}$ are the per-query retrieval and generation times, and $N$ is the number of queries. This metric reflects the amortized latency per query. These findings are further illustrated in Figures~\ref{fig:response_time_breakdown}, which visualize the total runtime and a 3D plot of average query latency

\subsubsection{Extra Storage Analysis}
Table~\ref{tab:extra-storage} presents the additional storage and expansion ratios for MiniRAG, Raptor, and our method on the NovelQA and Marathon datasets. MiniRAG leads to a 10.6x and 16.7x increase in data size, while Raptor results in a consistent 19.6x expansion across both datasets. In contrast, our method maintains a fixed 1.0x ratio, introducing no extra data.
The large overheads in MiniRAG and Raptor stem from storing dense similarity matrices and graph-based indexing structures, respectively. Our method avoids these components entirely, achieving zero storage overhead.

\begin{table}[h]
\centering
\scriptsize
\setlength{\tabcolsep}{4pt} 
\renewcommand{\arraystretch}{0.95} 
\begin{tabular}{l|c|c|c|c}
\toprule
\textbf{Dataset} & \textbf{Original Size} & \textbf{Method} & \textbf{Extra Data (MB)} & \textbf{Expansion Ratio} \\
\midrule
\multirow{3}{*}{NovelQA} 
    & \multirow{3}{*}{72.4 MB} 
    & MiniRAG   & 768.1 & 10.6x \\
    &           & Raptor    & 1421.2 & 19.6x \\
    &           & Ours      & 0.0    & 1.0x \\
\midrule
\multirow{3}{*}{Marathon} 
    & \multirow{3}{*}{146.1 MB} 
    & MiniRAG   & 2437.4 & 16.7x \\
    &           & Raptor    & 2862.6 & 19.6x \\
    &           & Ours      & 0.0    & 1.0x \\
\bottomrule
\end{tabular}
\caption{Extra data and expansion ratio on two datasets.}
\label{tab:extra-storage}
\end{table}

\section{Limitations}
Due to computational constraints, experiments were conducted with LLaMA models up to 70b parameters. While this offers a practical trade-off between capability and efficiency, it may understate the full potential of our retrieval framework. Future work could explore integration with larger backbones (e.g., GPT-4, Gemini 1.5, Claude Opus) to further assess scalability and generalization.
\section{Conclusion}
We introduced an embedding-free retrieval framework that fundamentally rethinks the architecture of Retrieval-Augmented Generation systems. By discarding vector embeddings and explicit graph construction, our method instead leverages the inferential and lexical capabilities of LLMs through an iterative, cognitively inspired process of search, evaluation, and refinement. This design enables real-time adaptability, transparent retrieval logic, and substantial gains in both efficiency and effectiveness.
Empirical evaluations on two challenging long-context QA benchmarks, NovelQA and Marathon, show that our approach not only outperforms state-of-the-art embedding-based systems in answer accuracy, but also achieves over an order of magnitude reduction in computation time and storage overhead. These improvements demonstrate that high-quality retrieval need not rely on opaque dense vectors or costly pre-processing pipelines.
\newpage
\bibliographystyle{plainnat} 
\bibliography{ref}      
\end{document}